\documentclass[runningheads]{llncs}
\usepackage{graphicx}
\usepackage{amsmath,amssymb} 
\usepackage{color}
\usepackage{setspace}
\usepackage{times}
\usepackage{epsfig}
\usepackage{booktabs}
\usepackage{bm}
\newcommand{\norm}[1]{\left\lVert#1\right\rVert}
\usepackage{latexsym}
\usepackage{listings}		
\usepackage{gensymb}
\usepackage[inline]{enumitem}
\usepackage{multirow}
\usepackage{bm}
\usepackage{breakurl}
\usepackage{url}
\usepackage[width=122mm,left=12mm,paperwidth=146mm,height=193mm,top=12mm,paperheight=217mm]{geometry}

\def\ECCV18SubNumber{}  

\begin{document}
\pagestyle{headings}
\mainmatter

\title{Improving Video Generation for Multi-functional Applications}

\authorrunning{Kratzwald et al.}

\institute{ETH Zurich}

\author{Bernhard Kratzwald, Zhiwu Huang, Danda Pani Paudel, Acharya Dinesh,\\Luc Van Gool\\
\tt \small{ \{kratzwab,acharyad\}@ethz.ch\\ \{zhiwu.huang,paudel,vangool\}@vision.ee.ethz.ch}}

\maketitle
\begin{abstract}

In this paper, we aim to improve the state-of-the-art video generative adversarial networks (GANs) with a view towards multi-functional applications. Our improved video GAN model does not separate foreground from background nor dynamic from static patterns, but learns to generate the entire video clip conjointly. Our model can thus be trained to generate---and learn from---a broad set of videos with no restriction. This is achieved by designing a robust one-stream video generation architecture with an extension of the state-of-the-art Wasserstein GAN framework that allows for better convergence. The experimental results show that our improved video GAN model outperforms state-of-the-art video generative models on multiple challenging datasets. Furthermore, we demonstrate the superiority of our model by successfully extending it to three challenging problems: video colorization, video inpainting, and future prediction. To the best of our knowledge, this is the first work using GANs to colorize and inpaint video clips.
\end{abstract}

\section{Introduction}

Viewed as a digital window into the real-life physics of our world, videos capture how objects behave, move, occlude, deform, and interact with each other. Furthermore, videos record how camera movements, scene depth or changing illumination influence a scene. Fully understanding their temporal and spatial dependencies is one of the core problems in computer vision. Teaching computers to model and interpret scene dynamics and dependencies occurring within videos is an essential step towards intelligent machines capable of interacting with their environment.

In contrast to the domain of images, the work on supervised and unsupervised learning from videos is still in its infancy. This can be attributed to the high-dimensional nature of videos. Performing large-scale supervised learning on video data requires prohibitively large amounts of labeled training samples. This can quickly become a bottleneck in supervised learning for video. The recent focus of research on videos has therefore shifted from supervised to unsupervised models. The near endless amount of unlabeled video data available on the Internet further encourages the choice of unsupervised methods~\cite{vondrick2017generating,walker2015dense,walker2016uncertain,walker2014patch}.   

State-of-the-art unsupervised video models are often designed to simplify the generation process by segmenting certain aspects of the video. Generative video models separate foreground from background~\cite{vondrick2016generating}, or dynamic from static patterns~\cite{TulyakovLYK17,saito2017temporal}. These are architectural choices that simplify and stabilize the generation process. On the other hand they often impose certain restrictions on the training data; e.g. \cite{vondrick2016generating} requires stable backgrounds and non-moving cameras. Video generation in a single stream avoids such simplifications but is inherently more difficult to achieve as low frequencies span both the temporal and spatial domain. The motivation of this work is to create a robust, universal and unrestricted generative framework that does not impose any preconditioning on the input videos while at the same time producing state-of-the-art quality videos. 

The task of generating videos is related to modeling and understanding the scene dynamics within them. For realistic video generation, it is essential to learn which objects move, how they move, and how they interact with each other, which \textit{vice versa} implies an understanding of real-world semantics. A model capable of understanding these semantics is ideally not restricted to the task of video generation but can also transfer this knowledge to a broad number of other applications. Important applications include action classification, object detection, segmentation, future prediction, colorization, and inpainting.

Our paper focuses both on the robustness of our generative video framework as well as on its application to three problems. First, we design a stable architecture with no prior constraints on the training data. More precisely, we design a one-stream generation framework that does not formally distinguish between foreground and background, allowing us to handle videos with moving backgrounds/cameras. Video generation in a single-stream is a fragile task, demanding a carefully selected architecture within a stable optimization framework. We accomplish this stability by exploiting state-of-the-art Wasserstein GAN frameworks in the context of video generation. In a second step, we demonstrate the applicability of our model by proposing a general multi-functional framework dedicated to specific applications. Our extension augments the generation model with an auxiliary encoder network and an application-specific loss function. With these modifications, we successfully conduct several experiments for unsupervised end-to-end training.

The two main contributions of this paper are as follows: (i) We propose iVGAN, a robust and unrestricted one-stream video generation framework. Our experiments show that iVGAN outperforms state-of-the-art generation frameworks on multiple challenging datasets. (ii) We demonstrate the utility of the multi-functional extension of iVGAN for three challenging problems: video colorization, video inpainting, and future prediction. To the best of our knowledge, this is the first work exploiting the advantages of GANs in the domains of video inpainting and video colorization.

\section{Related Work}
\textbf{Generative Adversarial Networks (GANs):}
GANs~\cite{goodfellow2014generative} have proven successful in the field of unsupervised learning. Generally, GANs consist of two neural networks: a generator network trained to generate samples and a discriminator network trained to distinguish between real samples drawn from the data distribution and fake samples produced by the generator. Both networks are trained in an adversarial fashion to improve each other. However, GANs are also known to be potentially unstable during training. To address this problem, Radford et al.~\cite{radford2015unsupervised} introduced a class of \emph{Deep Convolutional GANs} (DCGANs) that imposes empirical constraints on the network architecture. Salimans et al.~\cite{salimans2016improved} provide a set of tools to avoid instability and mode collapsing. Che et al.~\cite{CheLJBL16} use regularization methods for the objective to avoid the problem of missing modes. Arjovsky et al.~\cite{arjovsky2017wasserstein} suggest minimizing the Wasserstein-1 or Earth-Mover distance between generator and data distribution with theoretical reasoning.
In a follow-up paper, Gulrajani et al.~\cite{GulrajaniAADC17} propose an improved method for training the discriminator -- termed \emph{critic} by~\cite{arjovsky2017wasserstein} -- which behaves stably, even with deep ResNet architectures. GANs have mostly been investigated on images, showing significant success with tasks such as image generation~\cite{radford2015unsupervised,GulrajaniAADC17,karras2017progressive,DentonCSF15,ImKJM16}, image super-resolution~\cite{LedigTHCATTWS16}, style transfer~\cite{jurie1999new,ZhuPIE17}, and many others.

\textbf{Video Generation}: There has been little work on the topic of video generation so far~\cite{vondrick2016generating,TulyakovLYK17,saito2017temporal}. In particular, Vondrick et al.~\cite{vondrick2016generating} adapts the DCGAN model to generate videos, predict future frames and classify human actions. Their \emph{Video GAN} (VGAN) model suggests the usage of independent streams for generating foreground and background. The background is generated as an image and then replicated over time. A  jointly trained mask selects between foreground and background to generate videos. In order to encourage the network to use the background stream, a sparsity prior is added to the mask during learning. More recently, \emph{Temporal GAN} (TGAN)~\cite{saito2017temporal} deals with the instability in video generation by deploying a frame-wise generation model. A generative model for image generation is used to sample frames; a temporal generator preserves temporal consistency and controls this model. Tulyakov et al.~\cite{TulyakovLYK17} also adopted a two-stream generative model that produces dynamic motion vs. static content. In particular, the static part is modeled by a fixed Gaussian when generating individual frames within the same video clip, while a recurrent network that represents the dynamic patterns models the motion part. To deal with the instability of training GANs all three models separate integral parts of a video, as foreground from background or dynamic from static patterns. We argue that it is more natural to learn these patterns and their interference conjointly. Therefore, we propose a single-streamed but robust video generation architecture in Sec.~\ref{sec:methods}.

\textbf{Video Colorization:} Works on image and video colorization can be divided into two categories: interactive colorization that requires some kind of user input~\cite{chia2011semantic,huang2005adaptive,ironi2005colorization,levin2004colorization,luan2007natural,yatziv2006fast} and automatic methods~\cite{charpiat2008automatic,gupta2017learning,hertzmann2001image,liu2008intrinsic,welsh2002transferring,ZhangIE16}. Our approach belongs to the latter category. Most automatic methods come with restrictions that prevent them from working in general settings. For instance, \cite{liu2008intrinsic} requires colored pictures of a similar viewing angle and \cite{charpiat2008automatic} requires separate parameter tuning for every input picture. Methods such as~\cite{hertzmann2001image,welsh2002transferring} produce undesirable artifacts. In the video domain, methods such as~\cite{gupta2017learning} process each frame independently, which in turn leads to temporal inconsistencies. Recently, image colorization has been combined with GANs~\cite{koo2016automatic}, but no prior research on colorizing videos has been presented.  

\textbf{Video Inpainting:} Inpainting is a fairly well investigated problem in the image domain~\cite{bertalmio2000image,komodakis2006image,yang2016high}. For videos, it has been used to restore damage in vintage films~\cite{tang2011video}, to remove objects~\cite{granados2012background} or to restore error concealment~\cite{ebdelli2015video}. State-of-the-art frameworks like~\cite{le:hal-01492536} use complex algorithms involving optical flow computation; thus demanding an optimized version to run within a feasible amount of time. Recovering big areas of an image or a video, also called \emph{hole-filling}, is inherently a more difficult problem than the classical inpainting. Approaches like texture synthesis~\cite{barnes2009patchmatch,efros1999texture} or scene completion~\cite{hays2007scene} do not work for \emph{hole-filling}~\cite{PathakKDDE16}. While there has been some work on image inpainting with adversarial loss functions~\cite{PathakKDDE16}, we are not aware of any in the case of videos.

\textbf{Future Prediction:} Future prediction is the task of predicting the future frames for one/multiple given input frames. In contrast to video generation, future prediction is an elegant way of turning an unsupervised modeling problem into a supervised learning task by splitting videos into conditioning input and ground-truth future.
Our method builds upon recent future prediction work e.g.~\cite{walker2015dense,walker2016uncertain,walker2014patch,Chao2017,FinnGL16,FragkiadakiLM15,KalchbrennerOSD16,Luo2017,xue2016visual}, especially that using generative models and adversarial losses~\cite{vondrick2017generating,vondrick2016generating,MathieuCL15,ranzato2014video}.

\section{Our Model - iVGAN}
\label{sec:methods}
For robust video generation, we propose a simple yet tough to beat video generation model, called \emph{improved Video GAN} (iVGAN). Our model consists of a generator and a discriminator network in the GAN framework. Particularly, the designed generator $G: Z \rightarrow \mathcal{X}$  produces a video $\bm{x}$ from a low dimensional latent code $\bm{z}$. The proposed critic network $C:\mathcal{X} \rightarrow \mathbb{R}$ is optimized to distinguishing between real and fake samples and provides the generator updates with useful gradient information. 

Distinct from~\cite{vondrick2016generating}, we design the generation framework without any prior assumptions upon the nature of the data. Two-stream architectures generate the background as an image; thereby, limit the training data to videos with static backgrounds and non-moving cameras. It is thus essential that our generator is of one-stream, without separating back- and foreground. In contrast to~\cite{TulyakovLYK17,saito2017temporal} we use a simple but  effective architecture which learns spatial and temporal dependencies conjointly, rather than separating them into two networks. 

As studied in \cite{radford2015unsupervised,salimans2016improved,arjovsky2017wasserstein,GulrajaniAADC17} for image generation, it is non-trivial to train GAN models in a stable manner. Especially for video generation, it turns out to be much more challenging~\cite{saito2017temporal} as low frequencies also span the additional temporal domain. To address this problem, we generalize the state-of-the-art Wasserstein GAN to the context of video generation for more stable convergence. Formally, we place our network within the Wasserstein GAN framework~\cite{arjovsky2017wasserstein} optimizing 
\begin{equation}\label{eq:wgan}
\begin{split}
\min_{G}\max_{\|C\|_L\leq1}{V(G,C)} = \underset{\bm{x} \sim p_{data}(\bm{x})}{\mathbb{E}}[C(\bm{x})]- \underset{\bm{z} \sim p_z(\bm{z})}{\mathbb{E}}[C(G(\bm{z}))].
\end{split}
\end{equation}
In order to enforce the Lipschitz constraint on the critic function, we penalize its gradient-norm with respect to the input~\cite{GulrajaniAADC17}. For this purpose we evaluate the critic's gradient $\nabla_{\bm{\hat{x}}} C(\bm{\hat{x}})$ with respect to points sampled from a distribution over the input space $\bm{\hat{x}}\sim p_{\hat{x}}$, and penalize its squared distance from one via
\begin{equation}\label{eq:gpenalty}
\mathcal{L}_{GP}(C)=\underset
    {\bm{\hat{x}}\sim p_{\hat{x}}}
    {\mathbb{E}}
    \left[(\norm
        {\nabla_{\bm{\hat{x}}}C(\bm{\hat{x}})}_2-1
    )^2\right].
\end{equation}
The distribution $p_{\hat{x}}$ is defined by uniformly sampling on straight lines between points in the data distribution and points in the generator distribution. Hence, the final unconstrained objective is given by 
\begin{equation}\label{eq:genobj}
\min_{G}\max_{C} {V(G,C) + \lambda\,\mathcal{L}_{GP}(C) },
\end{equation}
where the hyperparameter $\lambda$ is used to balance the GAN objective with the gradient penalty.

\begin{figure}[t]
\begin{center}
\includegraphics[width=.75\linewidth]{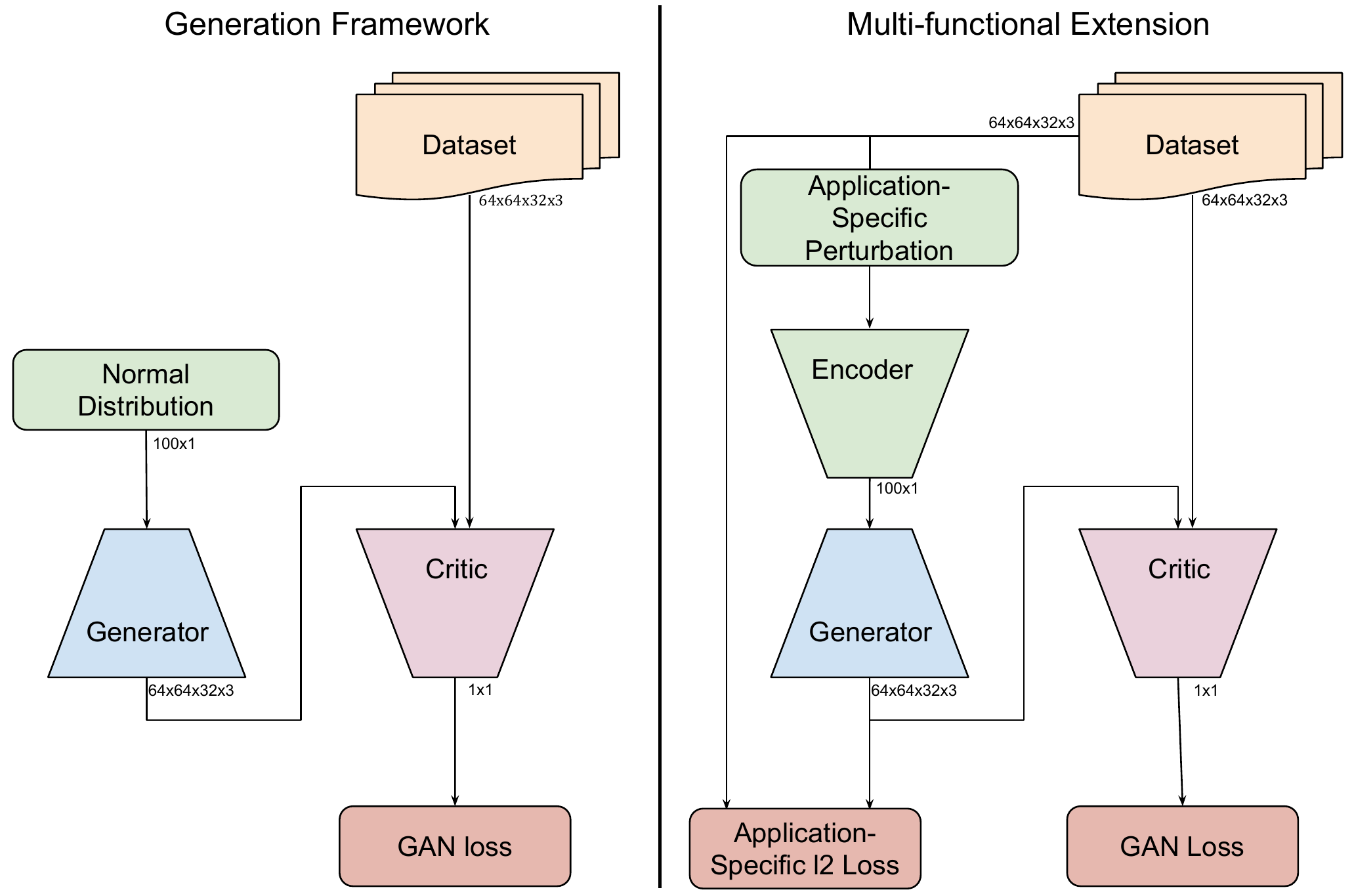}
\end{center}
 \caption{iVGAN video generation framework and its multi-functional extension}
\label{fig:generalization}
\end{figure}

\subsection{Generator Network}
The generator takes a latent code sampled from a 100-dimensional normal distribution $\bm{z} \sim \mathcal{N}(\bm{0},\bm{I})$ and produces an RGB video containing $32$ frames of $64\times64$ pixels. We use a linear up-sampling layer in the first step, producing a tensor of size $2\times4\times4\times512$. The linear block is followed by four convolutional blocks of spatio-temporal~\cite{ji20133d} and fractionally-strided~\cite{zeiler2010deconvolutional} convolutions. This combination has proven to be an efficient way to upsample, while preserving spatial and temporal invariances~\cite{vondrick2017generating,vondrick2016generating}. All convolutional layers utilize $4\times4\times4$ kernels, a stride of $2\times2\times2$, and add a bias to the output. We found the initialization of the convolutional weights essential for stable training and faster convergence. Inspired by the ResNet architecture~\cite{he2016deep} we initialize the kernels according to He et al.~\cite{he2015delving}. Similar to DCGAN~\cite{radford2015unsupervised}, all but the last layers are followed by a batch normalization layer~\cite{ioffe2015batch}. Batch normalization stabilizes the optimization by normalizing the inputs of a layer to zero mean and unit variance, which proved critical for deep generators in early training, preventing them from collapsing~\cite{radford2015unsupervised}.  

The first four blocks are followed by a ReLU non-linearity after the normalization layer, while the last layer uses a hyperbolic tangent function. This is beneficial to normalize the generated videos, identically to the videos in our dataset, within the range~$[-1,1]$.

\subsection{Critic Network}
The critic network maps an input video to a real-valued output. It is trained to distinguish between real and generated videos, while being constrained (Eqn.~\ref{eq:gpenalty}) to yield effective gradient information for generator updates.
 
The critic consists of five convolutional layers and is followed by an additional linear down-sampling layer. As in~\cite{vondrick2016generating}, we use spatio-temporal convolutions with $4\times4\times4$ kernels. Again we found the initialization of kernel weights important for stability and convergence during training and used the initializion following~\cite{he2015delving}. For more expressiveness, we add a trainable bias to the output. All convolutions include a stride of $2\times2\times2$ to enable efficient down-sampling of the high-dimensional inputs. 

Batch normalization correlates samples within a mini-batch by making the output for a given input $\bm{x}$ dependent on the other inputs $\bm{x'}$ within the same batch. A critic with batch normalization therefore maps a batch of inputs to a batch of outputs. On the other hand, in Eqn~\ref{eq:gpenalty}, we are penalizing the norm of the critic's gradient with respect to each input independently. For this reason, batch normalization is no longer valid in our theoretical setting. To resolve this issue, we use layer normalization~\cite{ba2016layer} following~\cite{GulrajaniAADC17}. Layer normalization works equivalent to batch normalization, but mean and standard deviation is calculated independently for every single sample $\bm{x_i}$ over the hidden layers. We found that layer normalization is not necessary for convergence, but essential if we optimize the generator with additional objectives, as described in the multi-functional extension in Sec.~\ref{ch:generalization}.

All but the last layer use a leaky ReLU~\cite{xu2015empirical} activation. We omit using a soft-max layer or any kind of activation in the final layer, since the critic is not trained to classify between real and fake samples, but rather trained to yield a good gradient information for generator updates.

\subsection{Learning and Parameter Configuration}
We optimize both networks using alternating stochastic gradient descent, more precisely we optimize the critic five times for every update step on the generator. The hyperparameter $\lambda$, controlling the trade-off between the GAN objective and the gradient penalty  (Eqn.~\ref{eq:genobj}), is set to $10$ as reported in~\cite{GulrajaniAADC17}. We use Adam~\cite{kingma2014adam} with initial hyperparameters $\alpha=0.0002$, $\beta_1=0.5$, $\beta_2=0.99$ and a batch size of $64$ which has proved to work best for us after testing various alternate settings. We divide the learning rate by two after visual convergence. We train our network from scratch which usually takes four to six days on a \emph{GeForce GTX TITAN X} GPU. The entire network is implemented in TensorFlow.

\section{Multi-functional Extension} 
\label{ch:generalization}
With a simple yet powerful modification, we extend our generation architecture to a multi-functional video processing framework. We choose three challenging applications to demonstrate the semantics our framework is capable of learning: (i)~to successfully colorize grayscale videos our network must learn temporally consistent color semantics;  meadows e.g. have to be painted in a shade of green which should stay consistent over time (ii)~inpainting, which is completing and repairing missing or damaged parts of a video, requires the network to learn spatial consistencies such as symmetries (iii)~future prediction conditioned on a single input frame is the toughest application and requires our model to learn and understand which objects are plausible to move how they do so. 

Fig.~\ref{fig:generalization} compares the generation framework architecture with its multi-functional extension. Similar to conditional GANs~\cite{mirza2014conditional}, the generator is no longer dependent on a randomly drawn latent code $\bm{z}$ but conditioned on additional application-specific information $\bm{y}$. A convolutional network $E : \mathcal{Y} \rightarrow \mathcal{Z}$ generates a latent code  $\bm{z}$ by encoding $\bm{y}$; which is in turn used to generate the desired video. To guide this generation we extend the framework by an additional application-specific loss $\mathcal{L}_{AP}$. 

The choice of $\bm{y}$ and the loss function depends on the application at hand. For video colorization we encode a grayscale video we wish to colorize and use the $\ell2$ loss between the generated and input video. For inpainting we condition on the damaged input clip and calculate the $\ell2$ loss between reconstruction and ground-truth. To predict future frames we encode a single input frame and apply the $\ell2$ loss between that frame and the first frame of the generated video. 

We jointly optimize for the GAN value function (Eqn.~\ref{eq:wgan}), the gradient penalty (Eqn.~\ref{eq:gpenalty}), and the new domain-specific loss $\mathcal{L}_{AP}$, using two hyperparameters $\lambda$ and $\nu$ to control the trade-off between them. To gain a deeper understanding of the interaction between GAN- and reconstruction loss, we conduct experiments with two variations of the colorization framework: In the \emph{unsupervised} setting the reconstruction loss is calculated in grayscale color space and does therefore not penalize wrong colorization, leaving the GAN-loss solely responsible for learning color semantics. In the \emph{supervised} setting on the other hand, the $\ell2$-loss is calculated in RGB color space and thus penalizes both wrong colorization and wrong structure. It remains unclear what role the GAN-loss takes in the latter setting. Following Zhao et al.~\cite{ZhaoML16} we argue in Sec.~\ref{ch:col} that the GAN-loss acts as a regularizer similar to a variational autoencoder; thus preventing the encoder-generator from learning a simple identity function.

\subsection{Learning and Parameter Configuration}
The encoder network consists of four strided convolutional layers, each of which is followed by a batch normalization layer and a ReLu activation function. We found it difficult to adjust the hyperparameter $\nu$ which controls the trade-off between the GAN loss and the domain-specific $\ell2$ loss. While the latter is per definition within the range $[0,1]$, the GAN loss is not bound as the critic output does not yield a probability anymore. We found it essential for a stable GAN loss to use layer normalization in the critic network; allowing us to monitor the losses and empirically set $\nu = 1000$.

\section{Experiments}
We evaluate our generation framework on multiple challenging datasets and compare our results with the two state-of-the-art video generation frameworks; namely the Video GAN (VGAN)~\cite{vondrick2016generating} and the Temporal GAN (TGAN) \cite{saito2017temporal} model. Other models such as~\cite{ranzato2014video} require supervision by one or more input frames and are hence excluded from our evaluation. For our multi-functional extension, we choose to colorize grayscale videos; inpaint damaged videos, and predict future frames from static images. Note that, for a better understanding, we also provide the readers with examples of animated generations and the source code for all our models in the supplementary material. 

\subsection{Datasets}
We used different datasets of unlabeled but filtered video clips, which have been extracted from high-resolution videos at a natural frame rate of $25$ frames per second. 

\textbf{Stabilized Videos:} This dataset\footnote{We downloaded the dataset from \url{http://carlvondrick.com/tinyvideo/}} was composed by~\cite{vondrick2016generating} and contains parts of the Yahoo Flickr Creative Commons Dataset~\cite{thomee2016yfcc100m}. The Places2 pre-trained model~\cite{zhou2014learning} has been used to filter the videos by scene category \emph{golf course}. All videos have been preprocessed to ensure a static background. Therefore, SIFT keypoints were extracted to estimate a homography between frames and minimize the background motion~\cite{vondrick2016generating}. The task of background stabilization may very often not be valid, forcing us to renounce a significant fraction of data. Discarding scenes with non-static background significantly restricts our goal of learning real-world semantics through unsupervised video understanding.

\textbf{Airplanes Dataset:} We compiled a second more challenging dataset of filtered, unlabeled and unprocessed video clips. Similar to the golf dataset videos are filtered by scene category, in this case \emph{airplanes}. Therefore, we collected videos from the YouTube-BoundingBoxes dataset~\cite{real2017youtube} which have been classified containing airplanes. No pre-processing of any kind has been applied to the data and the dataset thus contains static scenes as well as scenes with moving background or moving cameras.

\begin{figure}[t]
\begin{center}
\includegraphics[width=1\linewidth]{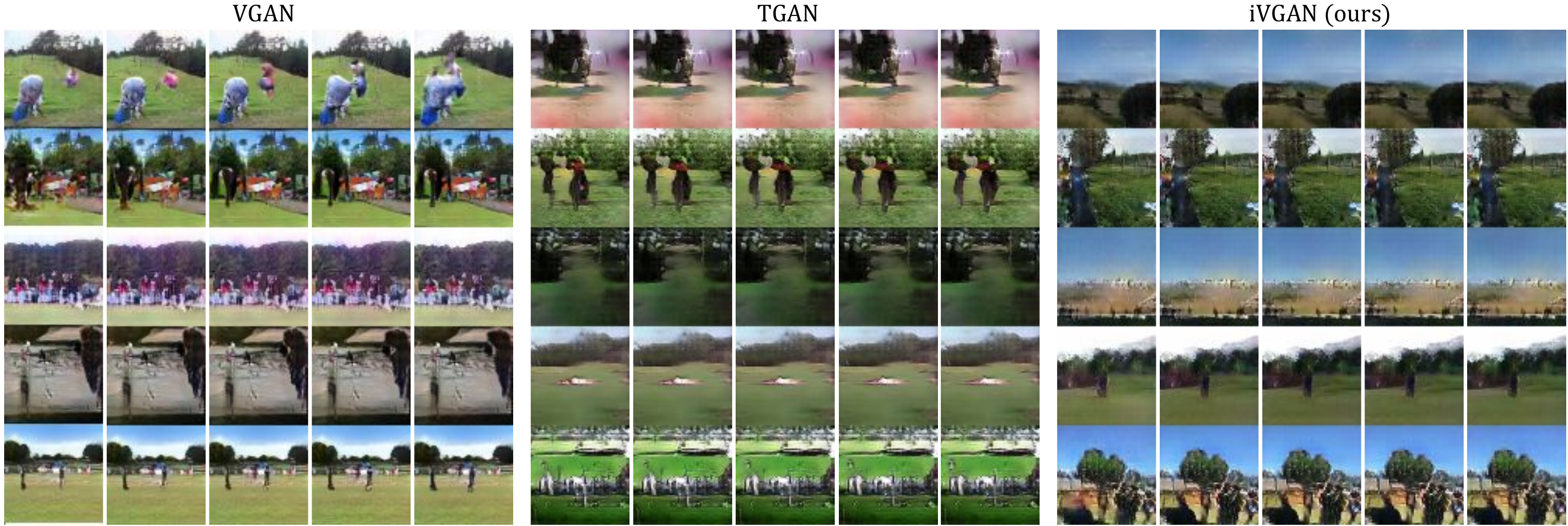}
\end{center}
 \caption{Video generation results on stabilized golf clips. \emph{Left}: Videos generated by the two-stream VGAN model. \emph{Middle}: Videos generated by the TGAN model. \emph{Right}: Videos generated by our one-stream iVGAN model}
\label{fig:golf}
\end{figure}

\begin{figure}[t]
\begin{center}
\includegraphics[width=1\linewidth]{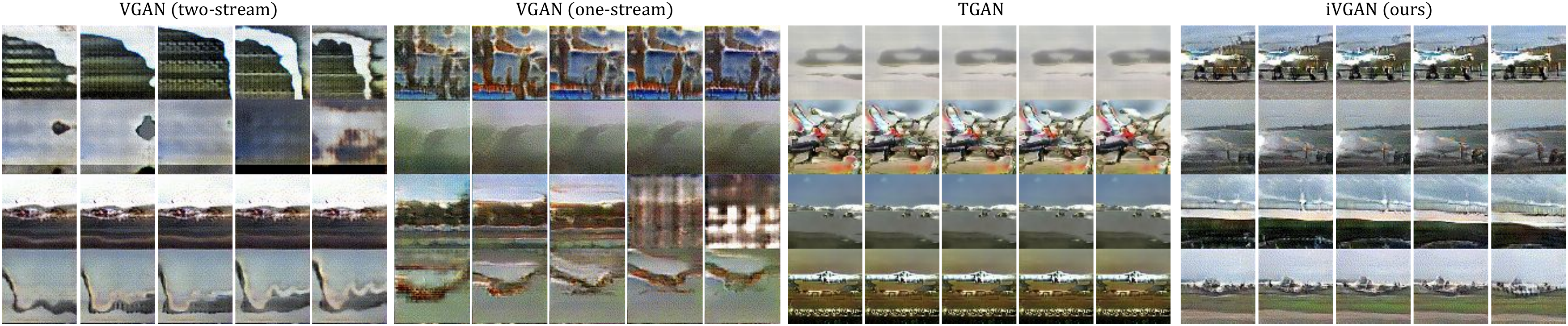}
\end{center}
 \caption{Video generation results on unstabilized airplane videos. Comparing videos generated using the one and two stream VGAN as well as the TGAN model, against our iVGAN framework}
\label{fig:airplanes}
\end{figure}

\subsection{Qualitative Evaluation}
Fig.~\ref{fig:golf} qualitatively compares results of the VGAN, TGAN and our iVGAN generator; where all three models were trained on the golf dataset. More animated samples are available in the supplementary material. There is no formal concept of foreground or background in the iVGAN model since the entire clip is generated in a single stream. Our model nonetheless naturally learns from the data to generate clips with a static background and moving foreground. Despite the fact that the background is not generated as an image (VGAN) it looks both sharp and realistic in the majority of samples. The foreground suffers from the same flaws as the VGAN and TGAN model: it is blurrier than the background, with people and other foreground objects turning into blobs. The network correctly learns which objects should move, and generates plausible motions. Some samples are close to reality, a fraction of samples collapse during training. Overall, the network learns correct semantics and produces scenes with a sharp and realistic looking background but blurry and only fairly realistic foreground-motion.

We conducted four independent experiments using the VGAN generator on the airplanes dataset, varying the learning rate between $0.00005$ and $0.0002$, and the sparsity penalty on the foreground mask between $0.1$ and $0.15$. In all runs, without exception, the generator collapsed and failed to produce any meaningful results. One might argue that it is unfair to evaluate a two-stream generation model, which assumes a static background, on a dataset violating this assumption. Therefore, we repeated a series of experiments using the one-streamed VGAN model, which does not separate foreground and background. A one-stream model should theoretically be powerful enough to converge on this dataset. Regardless of that, the one-stream version of VGAN collapsed as well in all experiments and failed to generate meaningful videos; indicating the difficulty of video generation with unstabilized videos. The more stable TGAN model does not collapse but fails to produce videos with moving backgrounds or camera motions.

Fig.~\ref{fig:airplanes} qualitatively compares generations from the two- and one-stream VGAN as well as the TGAN model against our iVGAN generator. Although the quality of our samples is lower compared to the stabilized golf videos, our generator did in no single experiment collapse. The iVGAN model -- unlike any other generative model -- produces both: videos with static background, as well as videos with moving background or camera motion. A fraction of the generated videos collapsed to meaningless colored noise, nonetheless. Nevertheless, it is clear that the network does learn important semantics since a significant number of videos shows blurry but realistic scenes, objects, and motions.
\begin{table}[t]
\centering
\singlespacing
\small
\caption{Quantitative Evaluation on Amazon Mechanical Turk: We show workers two pairs of videos and ask them which looks more realistic. We show the percentage of times workers prefer our model against real videos, VGAN and TGAN samples on two datastes}
\begin{tabular}{l|c}
\bfseries "Which video is more realistic?"    & \bfseries Percentage of Trials \\ \hline
Random Preference                  & 50                   \\ \hline
Prefer iVGAN over Real (Golf)      & 23.3                 \\
Prefer iVGAN over VGAN (Golf)      & 59.3                 \\
Prefer iVGAN over TGAN (Golf)      & 57.6                 \\ \hline
Prefer iVGAN over Real (Airplanes) & 15.4                 \\
Prefer iVGAN over TGAN (Airplanes) & 59.7                
\end{tabular}
\label{tab:eval-gen}
\end{table}

\subsection{Quantitative Results:} We used Amazon Mechanical Turk for a quantitative evaluation. Following~\cite{vondrick2016generating} we generated random samples from all three models as well as the original dataset. We showed workers a pair of videos drawn from different models and asked them: \emph{``Which video looks more realistic?''}. We paid workers one cent per comparison and required them to historically have a 95\% approval rating on Amazon MTurk. We aggregated results from more than 9000 opinions by 130 individual workers and show them in Tab.~\ref{tab:eval-gen}. Our results show that workers can clearly distinguish between real and fake videos; the distinction seems easier on the more challenging airplane dataset. Furthermore, workers asses that videos generated by our iVGAN model look significantly more realistic than those generated by the VGAN or TGAN model; hence, our iVGAN model clearly outperforms the state-of-the-art methods on both the golf and the airplane datasets. Since the VGAN model did not produce meaningful results on the airplane dataset we omitted the trivial comparison on this dataset.

\subsection{Colorization}
\label{ch:col}

\begin{figure}[t]
\begin{center} \includegraphics[width=.5\linewidth]{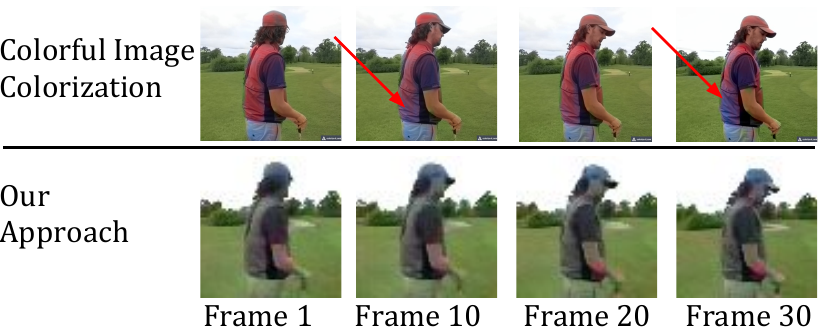}\end{center}
\caption{Color consistency over time with different colorization approaches. Red arrows mark spots where color is inconsistent over time}
\label{fig:colorization}
\end{figure}

Fig.~\ref{fig:colorization} qualitatively compares our framework with the state-of-the-art \emph{Colorful Image Colorization} (CIC) model~\cite{ZhangIE16}. The CIC model colorizes videos in their original resolution frame by frame. Our model, on the other hand, colorizes the entire clip at once but is restricted to in- and outputs of $64\times64$ pixels. Frame-wise colorization is known to suffer from temporal inconsistencies~\cite{gupta2017learning}. Fig.~\ref{fig:colorization} illustrates e.g. how the CIC colorized jacket changes its color over time while our colorization stays consistent. Our network overall learns correct color semantics: areas in the input are selected, ``classified'' and then painted accordingly. The sky e.g. is colorized in shades of blue or gray-white and trees are painted in a darker green than the grass. Therefore, we argue that the network not only selects the trees, but also recognizes (classifies) them as such, and paints them according to their class. The quality of the segmentation depends on the sharpness of the edges in the grayscale input. Colorized videos are blurrier compared to the grayscale input. This is mainly due to the fact that we do not keep the spatial resolution of the videos but encode them to a latent code, from which the colorized videos are then generated. Furthermore, using the mean squared error function to guide reconstructions is known to generate blurry results~\cite{MathieuCL15}. 

We evaluated the sharpness of the colorization quantitatively by the \emph{Peak Signal to Noise Ratio} (PSNR) in gray-space. PSNR correlates better with visual perception than the $\ell2$-loss. For the colorization quality we asked workers on Amazon MTurk to rate how realistic the video looks on a linear scale from 5 (very realistic) to 1 (very unrealistic). We generated random samples from each model and used random clips from the dataset as a reference value. The mean score for each model was calculated from more than 7000 ratings. We trained our models on 95\% of the golf dataset and evaluated them on 5\% hold-out data as well as on the out-of-domain airplane dataset. Notably even though we trained on stabilized video clips, our model is able to colorize clips with moving cameras and camera motion. The quantitative evaluation is shown in Tab.~\ref{tab:col-results}, animated results are available in the supplementary material.

\begin{table}[h]
\centering
\caption{Quantitative evaluation of video colorization and inpainting frameworks. Left: Average user rating of their realism from 1 (very unrealistic) to 5 (very realistic). Right: Peak signal to noise ratio between generated videos, and grayscale input (colorization) or ground-truth videos (inpainting)}
\singlespacing
\small
\begin{tabular}{lp{3cm}p{3cm}p{3cm}}
 \bfseries Model          &  \multicolumn{1}{c}{\bfseries MTurk} & \multicolumn{1}{c}{\bfseries PSNR}  &  \multicolumn{1}{c}{\bfseries PSNR} \\ 
                         &  \multicolumn{1}{c}{\bfseries~~average rating~~} & \multicolumn{1}{c}{\bfseries~~hold-out data~~} &  \multicolumn{1}{c}{\bfseries ~~out-of-domain data~~}\\
\midrule
\multicolumn{4}{c}{Video Colorization}                                                                             \\ \midrule
\textbf{supervised}     & \multicolumn{1}{c}{2.45}                    & \multicolumn{1}{c}{25.2 dB}                     & \multicolumn{1}{c}{23.4 dB}                          \\
\textbf{unsupervised}   & \multicolumn{1}{c}{2.95}                    & \multicolumn{1}{c}{25.6 dB}                     & \multicolumn{1}{c}{24.2 dB}                          \\ \midrule
\multicolumn{4}{c}{Video Inpainting}                                                                               \\ \midrule
\textbf{salt \& pepper} & \multicolumn{1}{c}{3.63}                    & \multicolumn{1}{c}{29.2 dB}                     & \multicolumn{1}{c}{25.4 dB}                          \\
\textbf{boxes (fixed)}  & \multicolumn{1}{c}{3.37}                    & \multicolumn{1}{c}{25.3 dB}                     & \multicolumn{1}{c}{22.9 dB}                          \\
\textbf{boxes (random)} & \multicolumn{1}{c}{3.43}                    & \multicolumn{1}{c}{24.7 dB}                     & \multicolumn{1}{c}{22.7 dB}                         
\end{tabular}
\label{tab:col-results}
\end{table}

To investigate the interplay between the GAN-loss and encoder-generator reconstruction loss we compare two variations of our model. As described in Sec.~\ref{ch:generalization}, the \emph{supervised} model calculates the reconstruction loss in RGB color space, while the \emph{unsupervised} model calculates the loss in grayscale color space. Our experiments indicate that the supervised colorization network, having a stronger objective, tends to overfit. Although they perform equally well on the training data, the unsupervised network outperforms the supervised network on hold-out and out-of-domain data as quantitatively shown in Tab.~\ref{tab:col-results}. The unsupervised model relies strongly on the GAN loss, which we argue -- following Zhao et al.~\cite{ZhaoML16} -- acts as a regularizer preventing the encoder-generator network from learning identity functions.

\begin{figure}[t]
\begin{center}
\includegraphics[width=1\linewidth]{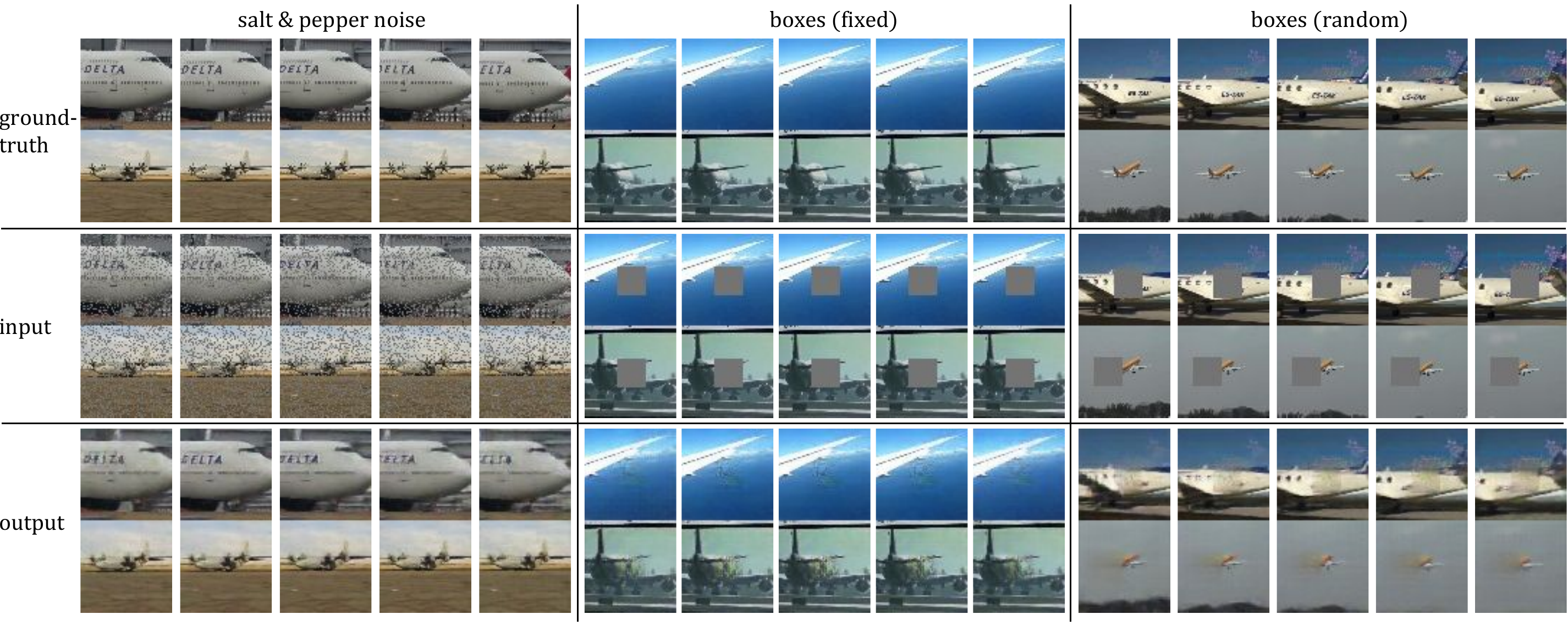}
\end{center}
 \caption{Comparison of ground-truth videos with the reconstructions of salt\&pepper noise, missing holes in the center and at random positions}
\label{fig:inpainting}
\end{figure}

\subsection{Inpainting}
We corrupt inputs in various ways and observe the reconstruction quality of our network: 25\%~salt and pepper noise, $20\times20$~pixel holes in the center of the clip, and $20\times20$~pixel holes at random positions. We trained our network on stabilized golf videos, and evaluate it on the unstabilized airplane dataset as shown in Fig.~\ref{fig:inpainting}.

Denoising salt and pepper corruptions is a well-studied problem, going back many years~\cite{chen1999tri}. State-of-the-art approaches operate on noise levels as high as 70\%~\cite{lu2012denoising}. The denoised reconstructions generated by our model are sharp and accurate. We can use our model -- which has been trained on stabilized videos --  to denoise clips with moving cameras or backgrounds, which would not be possible with a two-stream architecture. The reconstructed output is slightly blurrier than the ground-truth, which we attribute to the fact that we generate the entire video from a latent encoding and do not keep the undamaged parts of the input.

The task of hole-filling is more challenging since the reconstructions have to be consistent in both space and time. While we do not claim to compete with the state-of-the-art, we use it to illustrate that our network learns advanced spatial and temporal dependencies. For instance, in the second clip and second column of Fig.~\ref{fig:inpainting} we can see that, although the airplane's pitch elevator is mostly covered in the input, it is reconstructed almost perfectly and not split into two halves. This usually works best when the object covered is visible on more than one side of the box. We sometimes observe that such objects disappear although we could infer their existence from symmetry (e.g.  one airplane wing is covered and not reconstructed). Our model learns temporal dependencies, as objects which are covered in some---but not all frames---are reconstructed consistently over time. The overall quality does not suffer significantly when randomizing the locations of the boxes. 

Our quantitative evaluations results are shown in Tab.~\ref{tab:col-results}. We asked workers on Amazon MTurk to rate how realistic reconstructions look. Consistently with our quantitative findings, users rate the salt \& pepper reconstructions with a score of $3.63$ very high (real videos score $4.10$). The margin between boxes at fixed and random positions is very small and not significant. Furthermore, we calculate the peak signal to noise ratio between ground-truth videos and their reconstructed counterparts. Salt and pepper reconstructions achieve again the best score. The margin between boxes at fixed and boxes at random positions is too small to rank the models. All three models perform better on hold-out data than on the out-of-domain data.

\begin{figure}[h]
\begin{center}
\includegraphics[width=.5\linewidth]{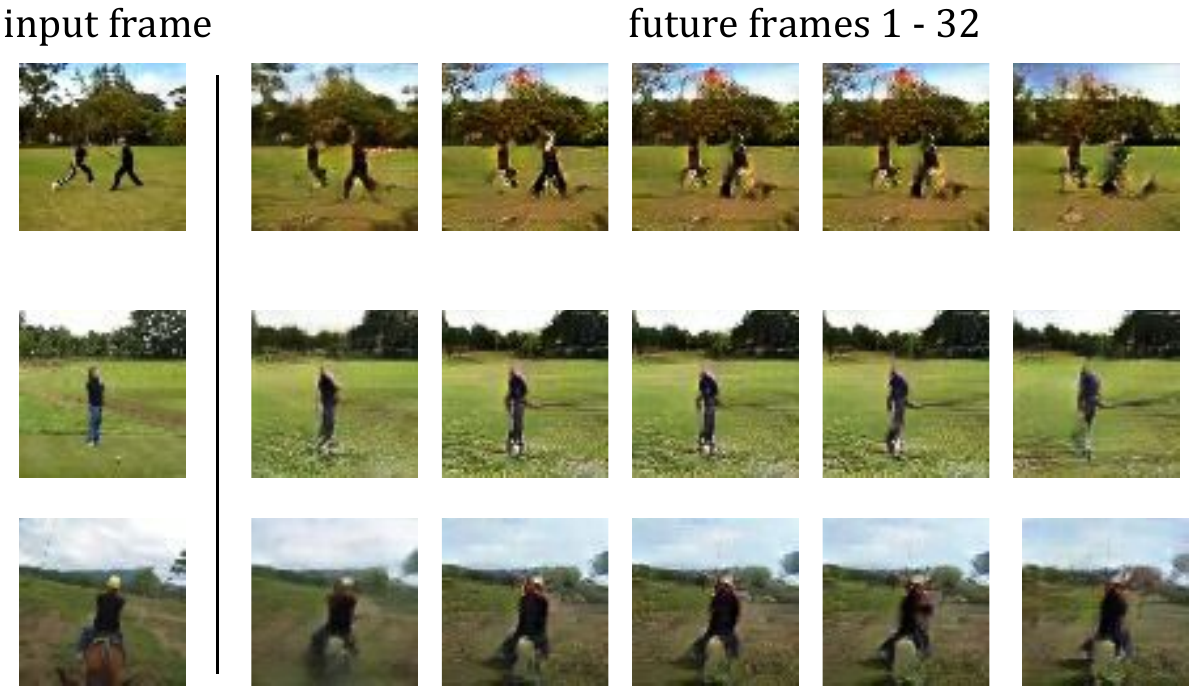}
\end{center}
 \caption{Future prediction results: Generated videos and the input frames the generations were conditioned on. The \emph{first} row shows two people who seem to fight. In person in the \emph{second} row seems to start walking. The Person in the \emph{third} row rides a horse; the horse is dropped in the future frames but the person moves}
\label{fig:future}
\end{figure}

\subsection{Future Prediction}
We qualitatively show results of our future prediction network in Fig.~\ref{fig:future}. Future frames are blurrier, compared to the inpainting and colorization results, which we attribute to the fact that the reconstruction loss only guides the first frame of the generated clip -- not the entire clip.

Although in many cases the network fails to generate a realistic future, it often learns which objects should move and generates fairly plausible motions. Since we use only one frame guiding the generation and omit to use the ground-truth future, these semantics are solely learned by the adversarial loss function. We emphasize that, to the best of our knowledge, this work and~\cite{vondrick2016generating} are the only two approaches using a single input frame to generate multiple future frames. We suffer from the same problems as~\cite{vondrick2016generating}, such as hallucinating or omitting objects. For example, the horse in the bottom-most clip in Fig.~\ref{fig:future} is dropped in future frames. Unsupervised future prediction from a single frame is a notoriously hard task. Nonetheless, our network learns which objects are likely to move, and to generate fairly plausible motions. 

\section{Conclusion and Outlook}

This paper proposed a robust video generation model that generalizes the state-of-the-art Wasserstein GAN technique to videos, by designing a new one-stream generative model. Our extensive qualitative and quantitative evaluations show that our stable one-stream architecture outperforms the Video GAN and Temporal GAN models on multiple challenging datasets. Further, we have verified that one-stream video generation can work within a suitable framework and stable architecture. The proposed iVGAN model does not need to distinguish between foreground and background or dynamic and static patterns and is the only architecture able to generate videos with moving camera/background, as well as those with a static background. Although our architecture does not explicitly model the fact that our world is stationary, it correctly learns which objects might plausibly move and how. 

Additionally, dropping the assumption of a static background frees our model to handle data that is not background-stabilized, thus significantly broadening its applicability. We emphasized the superiority of our model by demonstrating that our proposed multi-functional extension is applicable to several distinct applications, each of them requiring our network to learn different semantics. Our video colorization experiments indicate that the model is able to select individual parts of a scene, recognize them, and paint them accordingly. The inpainting experiments show that our model is able to learn and recover important temporal and spatial dependencies by filling the damaged holes consistently, in both space and time. We trained our models on stabilized input frames in both applications and successfully applied them to unprocessed videos. A two-stream model would by design not be able to colorize or inpaint clips exhibiting background or camera motion.

Although unsupervised understanding of videos is still in its infancy, we have presented a more general and robust video generation model that can be used as a multi-functional framework. Nevertheless, we believe that the quality of the generated videos can be further improved by using deeper architectures like ResNet~\cite{he2016deep} or DenseNet~\cite{huang2016densely}, or by employing recent progressive growing techniques of GANs~\cite{karras2017progressive}.

\end{document}